\pdfoutput=1
\documentclass[11pt,a4paper]{article}
\usepackage[utf8]{inputenc}
\usepackage[T1]{fontenc}
\usepackage{lmodern}
\usepackage[margin=2.4cm]{geometry}
\usepackage{amsmath,amssymb}
\usepackage{graphicx}
\usepackage{booktabs}
\usepackage{array}
\usepackage{tabularx}
\usepackage[dvipsnames]{xcolor}
\usepackage[numbers,square,sort&compress]{natbib}
\usepackage{authblk}
\usepackage[font=small,labelfont=bf,labelsep=period]{caption}
\usepackage[hidelinks]{hyperref}
\usepackage{pdfpages}

\title{\textbf{Where slide-level foundation-model classifiers fail on atypical breast lesions}}

\author[1,*]{Nerea Hernandez}
\author[1]{Luis J. Herrera}
\author[2]{Cesar Ramirez}
\author[3]{Olga Valenzuela}
\author[1]{Ignacio Rojas}
\author[1]{Francisco M. Ortuño}
\affil[1]{Department of Computer Engineering, Automation and Robotics, CITIC-UGR, University of Granada, Granada, 18071, Spain}
\affil[2]{Hospital Universitario Clínico San Cecilio, Granada, 18016, Spain}
\affil[3]{Department of Applied Mathematics, University of Granada, Granada, 18071, Spain}
\affil[*]{Corresponding author: nereahc@ugr.es}
\date{}

\begin{document}
\maketitle

\begin{abstract}

\noindent Whether a ductal breast lesion is called atypical hyperplasia or low-grade carcinoma in situ increasingly determines surgery. However, slide-level classifiers built on pathology foundation models (PFMs) perform worst on these lesions. Using four PFMs, cross-validation on BRACS and preregistered hypotheses, we studied where the classification of atypia fails. Because slides are labelled with their most severe lesion, only 39\% of the slides with annotated atypical ductal hyperplasia (ADH) were labelled ADH. Attention ranked the lesion high but rarely first, and showing the model only the lesion helped little, whereas classifiers trained per decision with lesion-level labels improved all four atypical boundaries. ADH regions inserted into benign slides raised the probability of malignancy more than that of atypia. In a hospital with pathologist-labelled single-subtype slides and in HISTAI, benign slides were scored closer to lesions and triage specificity fell from 0.48--0.53 to 0.00--0.07 and 0.22.

\end{abstract}

\section*{Introduction}

The boundaries between benign, atypical and in situ ductal lesions of the breast determine treatment. Usual ductal hyperplasia (UDH) is followed up. Atypical ductal hyperplasia (ADH), and often flat epithelial atypia (FEA), lead to excision or to risk-reducing measures. Ductal carcinoma in situ (DCIS) has traditionally been treated as early cancer, but this boundary is now moving. In the COMET randomised trial, active monitoring of low-risk DCIS (hormone receptor-positive, grade 1 or 2) was not inferior to surgery, with or without radiotherapy, in the two-year rate of invasive cancer~\citep{hwang_comet_2025}. Therefore, whether a lesion is classified as UDH, ADH, or low- or high-grade DCIS increasingly decides whether a woman undergoes surgery.

These are also the least reproducible diagnoses in breast pathology. In the B-Path study, pathologists agreed with a reference diagnosis of atypia in about half of the cases, against more than nine in ten for invasive carcinoma~\citep{elmore_concordance_2015}. Agreement among UDH, ADH and DCIS was fair ($\kappa=0.34$)~\citep{jain_interobserver_2011}, experts agreed only fairly on atypia in columnar cell lesions~\citep{el_sheikh_diagnostic_2025}, and agreement on the grade of DCIS was moderate ($\kappa=0.50$)~\citep{vanseijen_grading_2021}. Part of this disagreement comes from the disease itself. FEA, ADH and low-grade DCIS form a biological, genetic and morphological continuum rather than discrete entities~\citep{collins_precursor_2018, WHO_2019_Breast}, and atypical lesions overlap histologically with benign mimics and with malignant precursors~\citep{renshaw_long_2016, hartmann_atypical_2015}. FEA differs from columnar cell change mainly in subtle cytology, whereas ADH shares its cytology and architecture with low-grade DCIS and is separated from it by extent, about 2~mm or two duct spaces~\citep{pinder_ellis_2003, vandenbussche_borderline_2013, tozbikian_adh_2017, korie_interobserver_2025}. This extent has to be measured across the slide and cannot be read from a single field.

Computational pathology has obtained better results for the least and the most severe lesions than for the intermediate ones. Carcinoma is detected reliably on whole slides~\citep{campanella_clinical_2025, ehteshami_bejnordi_diagnostic_2017}, DCIS included, with areas under the curve near 0.98 in a multi-site validation~\citep{sandbank_validation_2022}. Within DCIS, Dong et al.\ separated low from high grade with nuclear features~\citep{dong_computational_2014}, and the grader of Wetstein et al.\ agreed with expert observers as well as they agreed with each other~\citep{wetstein_dcis_2021}. Intermediate lesions have been more difficult. A five-class whole-slide classifier that included ADH reached 55\% accuracy, no different from that of 45 pathologists~\citep{gecer2018}, and more recent systems, one of them in clinical routine, group ADH and low-grade DCIS into a single class~\citep{sandbank_validation_2022, yoo_trpv4_2026}. In the external cohort of Yoo et al., H\&E models of this spectrum reached a macro-F1 of only 0.43~\citep{yoo_trpv4_2026}.

Pathology foundation models (PFMs), pre-trained on millions of histopathology images~\citep{vorontsov_foundation_2024, ochi_pathology_2025, chen_towards_2024, zimmermann_virchow2_2024, xu_whole-slide_2024}, raised the hope that larger representations would close this gap. However, benchmarks find even the strongest PFMs variable on pre-invasive lesions~\citep{ma_pathbench_2025}, and in a comparison of 32 foundation models, neither model size nor the scale of pre-training predicted downstream performance~\citep{bareja_benchmark_2026}. On BRACS, the public reference dataset for atypia~\citep{brancati_bracs_2022}, new encoders and aggregators~\citep{shao_transmil_2021, zhang_2dmamba_2025, zheng_m3amba_2025, zhao_mexd_2025} have raised aggregate AUCs. None of these studies reports slide-level performance for ADH or FEA, and the same column name is used for micro-, macro- and weighted F1 (Supplementary Note S16). With strong frozen features, the encoder matters more than the aggregator~\citep{shao_multiple_2025} and spatial context adds little~\citep{li_spatial_2026}. To our knowledge, no study has measured at which step a slide-level PFM classifier fails on atypia.

Two properties of the problem are rarely made explicit. The first is how the label is generated. In BRACS, the slide label is the most aggressive subtype annotated in the slide, and the annotated regions are representative rather than exhaustive~\citep{brancati_bracs_2022}, so a slide with ADH next to DCIS is labelled as DCIS and its ADH does not appear in the label. The second is where the evidence is. Under weak supervision, an attention-based aggregator~\citep{ilse_attention-based_2018, lu_data-efficient_2021} has to locate the lesion that defines the label and also assign it the correct subtype. A failure on atypia can therefore arise at either step, in the label, or at the deployment site, where PFMs encode the acquisition centre~\citep{komen_towards_2026} and the optimal decision thresholds change~\citep{pignet_robust_2025}.

In this work, instead of proposing a new classifier for atypia, we studied at which step its classification fails. We divided the problem into four steps (the label, the lesion, each atypical boundary and the deployment site; Fig.~\ref{fig:design}) and tested each of them with hypotheses written before the results existed, including causal probes in the space of patch embeddings~\citep{kaczmarzyk_explainable_2024}. One of the two external cohorts consists of slides that each contain a single subtype labelled by a pathologist, which separates the effect of the label from that of the model. We show that most of the annotated atypia does not appear in the slide labels, that the slide model fails mainly when it has to classify the lesion rather than when it has to locate it, and that at new sites the triage threshold loses specificity because benign slides are scored closer to lesions as a result of differences in image acquisition. We release the folds, the per-boundary evaluation and the probes so that other models can be tested in the same way.

\section*{Results}

\subsection*{Study design}

\begin{figure*}[t]
\centering
\includegraphics[width=\textwidth]{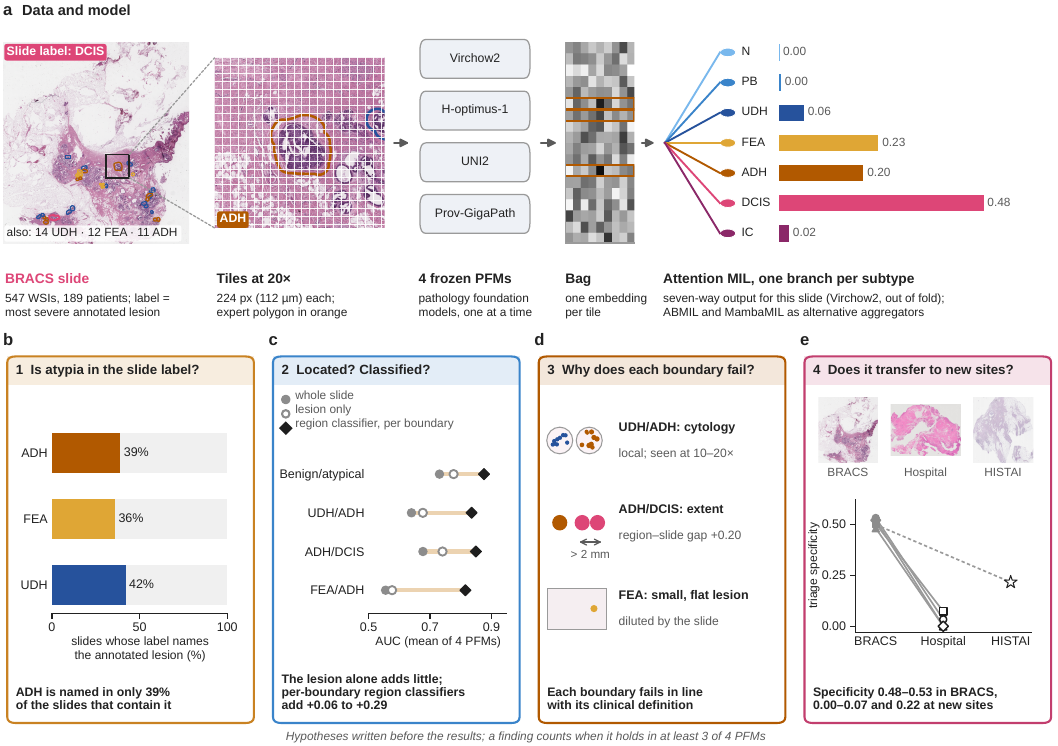}
\caption{\textbf{Study overview.} \textbf{a}, Data and model, shown for one BRACS slide. The slide is labelled DCIS after its most severe lesion, although UDH, FEA and ADH regions are also annotated (coloured outlines, expert polygons). The tissue is cut into 224-pixel tiles at 20$\times$ (inset, around an annotated ADH region), each tile is encoded by one of four frozen pathology foundation models (PFMs), and the bag of tile embeddings is read by an attention-based multiple instance learning (MIL) model with one branch per subtype; right, its seven-way output for this slide (Virchow2, out of fold, mean of three repetitions). \textbf{b--e}, The four questions of the study and their main result. \textbf{b}, Percentage of BRACS slides with an annotated UDH, FEA or ADH region whose label names that lesion (Fig.~\ref{fig:label}). \textbf{c}, AUC of the four atypical boundaries with the whole slide, with only the lesion, and with a region classifier trained per boundary on lesion-level labels; mean of four PFMs (Fig.~\ref{fig:map}). \textbf{d}, The criterion that defines each atypical boundary and how it fails (Figs.~\ref{fig:probes} and~\ref{fig:boundaries}). \textbf{e}, One slide from each centre, and the specificity of the triage threshold fixed in BRACS cross-validation, in BRACS and at the two new sites; markers as in Fig.~\ref{fig:deploy}b.}
\label{fig:design}
\end{figure*}

We analysed the 547 whole-slide images (WSIs) of BRACS and two external cohorts (Fig.~\ref{fig:design}). The first comprised 118 WSIs from a second hospital, each selected by a pathologist to contain a single subtype, and the second 517 cases of the public HISTAI collection, with labels taken from their pathology reports (Methods). Features were extracted at 20$\times$ with four PFMs (Virchow2, H-optimus-1, UNI2 and Prov-GigaPath), and an attention-based multiple instance learning (MIL) model with one branch per subtype was trained on the seven subtypes. It was evaluated with patient-grouped five-fold cross-validation repeated three times, and the external cohorts were scored without adaptation. Region-level analyses used the expert polygons of the 387 annotated slides, which never entered training. Every confirmatory hypothesis was written before the corresponding result existed and was considered supported when it held in at least three of the four PFMs. The hypotheses and their outcomes are listed in Supplementary Note S15, and later analyses are labelled exploratory. Intervals are 95\% patient-bootstrap intervals, and ranges go from the minimum to the maximum across the four PFMs.

On the three-class task the model was comparable to published results, with a cross-validated macro AUC of 0.86--0.88 (Table~\ref{tab:rendimiento}). However, its atypical F1 was only 0.40--0.43, and the seven-way F1 of ADH 0.15--0.23. In the hospital the three-class AUC changed little (0.82--0.86) while the atypical F1 collapsed. Aggregate metrics therefore do not reveal the poor results on atypia, and the rest of this section examines where they arise.

\begin{table*}[t]
\centering
\caption{\textbf{Slide-level performance.} Rows are the seven-way attention model (ce7) used throughout, except the official-split row ``ce3'', the same model and recipe trained on the three superclasses, as in the literature. Three-class metrics as reported in the literature (Supplementary Note S16): accuracy, macro F1 and one-versus-rest macro AUC of the superclass probabilities. AUC AT and F1 AT, atypical versus the rest; F1 FEA and F1 ADH, per-subtype F1 of the seven-way prediction; RPS7, ranked probability score over the seven subtypes (lower is better). Mean $\pm$ s.d.\ over repetitions (cross-validation and external cohorts) or seeds (official split, run with Virchow2 only). F1 FEA is not evaluated externally (one FEA slide); subtype metrics do not exist for ce3.}
\label{tab:rendimiento}
\resizebox{\textwidth}{!}{
\begin{tabular}{@{}lcccccccc@{}}
\toprule
\textbf{PFM} & \textbf{Acc.} & \textbf{Macro F1} & \textbf{AUC} & \textbf{AUC AT} & \textbf{F1 AT} & \textbf{F1 FEA} & \textbf{F1 ADH} & \textbf{RPS7} \\ \cmidrule(lr){2-4} \cmidrule(lr){5-6} \cmidrule(lr){7-8} & \multicolumn{3}{c}{three classes} & \multicolumn{2}{c}{atypical} & \multicolumn{2}{c}{seven classes} & \\ \midrule
\multicolumn{9}{@{}l}{\textit{BRACS, patient-grouped CV (3 repetitions)}} \\
Virchow2 & 0.77 $\pm$ 0.02 & 0.70 $\pm$ 0.02 & 0.88 $\pm$ 0.01 & 0.78 $\pm$ 0.02 & 0.43 $\pm$ 0.04 & 0.19 $\pm$ 0.07 & 0.23 $\pm$ 0.04 & 0.096 $\pm$ 0.009 \\
H-optimus-1 & 0.74 $\pm$ 0.01 & 0.68 $\pm$ 0.02 & 0.88 $\pm$ 0.00 & 0.78 $\pm$ 0.01 & 0.42 $\pm$ 0.05 & 0.20 $\pm$ 0.06 & 0.15 $\pm$ 0.03 & 0.103 $\pm$ 0.002 \\
Prov-GigaPath & 0.74 $\pm$ 0.01 & 0.67 $\pm$ 0.01 & 0.86 $\pm$ 0.01 & 0.77 $\pm$ 0.01 & 0.40 $\pm$ 0.03 & 0.17 $\pm$ 0.08 & 0.18 $\pm$ 0.02 & 0.108 $\pm$ 0.011 \\
UNI2 & 0.76 $\pm$ 0.01 & 0.69 $\pm$ 0.02 & 0.87 $\pm$ 0.01 & 0.77 $\pm$ 0.02 & 0.40 $\pm$ 0.02 & 0.25 $\pm$ 0.04 & 0.18 $\pm$ 0.02 & 0.098 $\pm$ 0.002 \\
\addlinespace
\multicolumn{9}{@{}l}{\textit{BRACS, official split (10 seeds)}} \\
Virchow2, ce7 & 0.69 $\pm$ 0.03 & 0.64 $\pm$ 0.06 & 0.87 $\pm$ 0.02 & 0.79 $\pm$ 0.05 & 0.38 $\pm$ 0.16 & 0.10 $\pm$ 0.11 & 0.12 $\pm$ 0.12 & 0.104 $\pm$ 0.009 \\
Virchow2, ce3 (three classes) & 0.73 $\pm$ 0.02 & 0.71 $\pm$ 0.03 & 0.89 $\pm$ 0.01 & 0.82 $\pm$ 0.02 & 0.53 $\pm$ 0.07 & --- & --- & --- \\
\addlinespace
\multicolumn{9}{@{}l}{\textit{External cohort 1: hospital (5-model ensemble, 3 repetitions)}} \\
Virchow2 & 0.80 $\pm$ 0.02 & 0.66 $\pm$ 0.05 & 0.86 $\pm$ 0.01 & 0.76 $\pm$ 0.01 & 0.28 $\pm$ 0.12 & --- & 0.06 $\pm$ 0.10 & 0.101 $\pm$ 0.001 \\
H-optimus-1 & 0.79 $\pm$ 0.03 & 0.61 $\pm$ 0.07 & 0.86 $\pm$ 0.01 & 0.75 $\pm$ 0.03 & 0.14 $\pm$ 0.16 & --- & 0.10 $\pm$ 0.08 & 0.111 $\pm$ 0.002 \\
Prov-GigaPath & 0.68 $\pm$ 0.09 & 0.56 $\pm$ 0.01 & 0.85 $\pm$ 0.02 & 0.72 $\pm$ 0.05 & 0.20 $\pm$ 0.18 & --- & 0.14 $\pm$ 0.12 & 0.122 $\pm$ 0.015 \\
UNI2 & 0.64 $\pm$ 0.06 & 0.47 $\pm$ 0.08 & 0.82 $\pm$ 0.00 & 0.65 $\pm$ 0.01 & 0.05 $\pm$ 0.09 & --- & 0.00 $\pm$ 0.00 & 0.137 $\pm$ 0.008 \\
\addlinespace
\multicolumn{9}{@{}l}{\textit{External cohort 2: HISTAI-breast (case-level labels, 5-model ensemble, 3 repetitions)}} \\
Virchow2 & 0.73 $\pm$ 0.03 & 0.55 $\pm$ 0.03 & 0.78 $\pm$ 0.00 & 0.67 $\pm$ 0.00 & 0.16 $\pm$ 0.07 & --- & 0.09 $\pm$ 0.08 & 0.152 $\pm$ 0.001 \\
\bottomrule
\end{tabular}
}
\end{table*}

\subsection*{Most annotated atypia does not appear in the slide label}

\begin{figure*}[p]
\centering
\includegraphics[width=\textwidth]{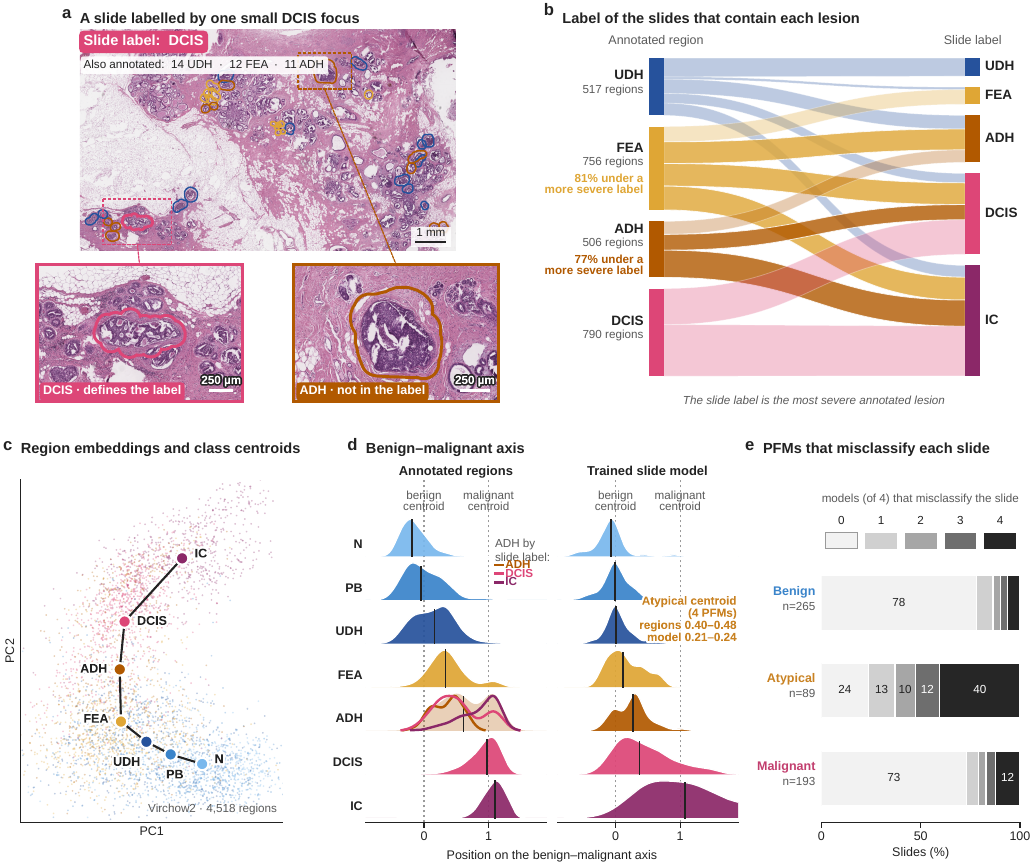}
\caption{\textbf{Most annotated atypia is not reflected in the slide label.} \textbf{a}, A BRACS slide labelled DCIS because of one small focus (left inset), although 14 UDH, 12 FEA and 11 ADH regions are also annotated (right inset, one of the ADH regions); outlines, expert polygons coloured by subtype. Scale bars, 1~mm (overview) and 250~$\mu$m (insets). \textbf{b}, Annotated regions of each lesion (left) by the label of the slide that contains them (right). Because the label is the most severe annotated lesion, flows can only go to an equal or more severe label; percentages, regions under a more severe label. \textbf{c}, Region embeddings (Virchow2; mean patch embedding inside each polygon) on the first two principal components, for display only; large markers, class centroids joined in canonical order. The order is established on the full embedding by the shortest path through the centroids (Methods). \textbf{d}, Position on the axis from the benign centroid (0) to the malignant centroid (1) of every annotated region (left) and of every slide in the embedding of the trained slide model (right; first repetition), by subtype; kernel densities, with medians as vertical lines. For ADH regions the shaded profile is split by the label of the slide that contains them (lines). Text, position of the atypical centroid in the four PFMs. \textbf{e}, Number of PFMs, of four, that misclassify each slide's superclass, by true superclass (patient-grouped cross-validation; out-of-fold probabilities averaged over three repetitions); numbers, percentage of slides.}
\label{fig:label}
\end{figure*}

In BRACS, 63\% of the annotated slides contain more than one subtype, and the label keeps only the most severe one (Fig.~\ref{fig:label}a). The intermediate lesions were the most affected. Only 39\% of the slides with annotated ADH were labelled as ADH, and 77\% of the ADH polygons lay in slides labelled with a more severe class (Fig.~\ref{fig:label}b). Conversely, 69\% of the DCIS slides also contained ADH or FEA. Most of the atypical tissue that a slide model sees during training therefore carries the label of another lesion.

This mixing also affects how the model is evaluated. When ADH slides were compared only with the DCIS slides that had no ADH or FEA annotated, the AUC of ADH versus DCIS was 0.76--0.79, against 0.61--0.65 for the DCIS slides that had them (exploratory; 16 and 35 DCIS slides, intervals including zero in 4/4 PFMs). In the hospital, where each slide contains a single subtype, the same decision reached 0.92--0.96 (see below).

The region embeddings kept the clinical order of the subtypes. The shortest path through the seven class centroids followed the canonical order from normal tissue to invasive carcinoma in 4/4 PFMs, and the atypical centroid lay about halfway between benign and malignant (Fig.~\ref{fig:label}c). The trained slide model preserved this order only partially and placed atypia closer to benign (Fig.~\ref{fig:label}d). The ADH regions themselves did not form a single population. Their distribution was bimodal, and the upper mode, located where DCIS lies, contained 64--76\% of the ADH regions of invasive carcinoma (IC) slides but only 1--10\% of those of ADH slides. These regions were not fragments of DCIS by the extent criterion, since they were as large as the ADH of ADH slides and lay no closer to carcinoma (preregistered; Supplementary Note S13). This bimodality was found after the confirmatory analyses and rests on 21 patients.

Finally, the four PFMs failed on the same slides. All four misclassified 40\% of the atypical slides, compared with 6\% of the benign and 12\% of the malignant slides (Fig.~\ref{fig:label}e), so combining them is unlikely to help with atypia.

\subsection*{The slide model locates part of the lesion but does not classify it}

\begin{figure*}[p]
\centering
\includegraphics[width=\textwidth]{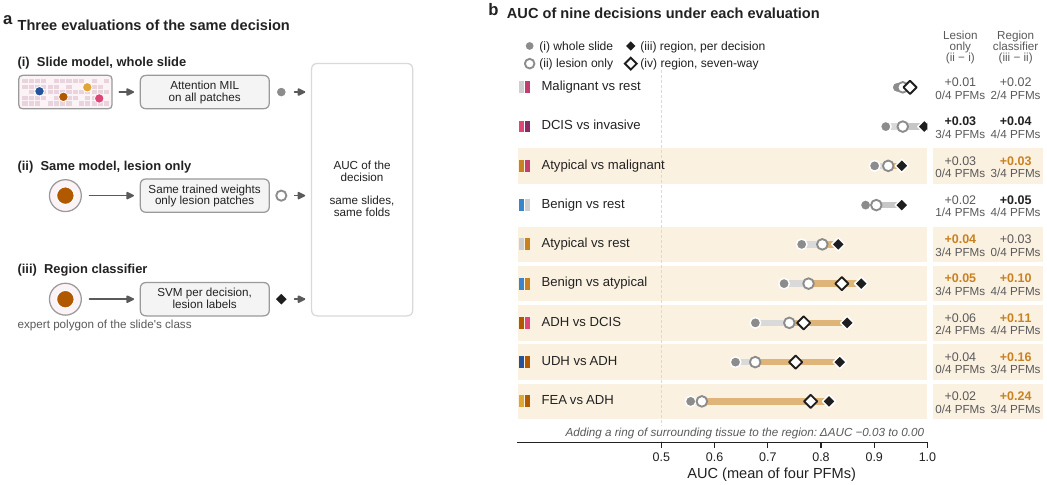}
\caption{\textbf{Performance by clinical decision and effect of localising the lesion.} \textbf{a}, Three evaluations of the same decision, scored on the same annotated slides and folds: (i) the slide model (attention MIL over every patch of the slide), which has to find the lesion itself; (ii) the same trained model, without retraining, given only the patches inside the expert polygons of the slide's class; and (iii) a region classifier (support vector machine on the mean embedding of each polygon) trained for each decision with lesion-level labels. Only annotated slides enter, so evaluation (i) differs slightly from the all-slide AUCs in the text. \textbf{b}, AUC of nine clinical decisions under the three evaluations and, for the five preregistered decisions, (iv) a single seven-way classifier trained on the same regions with lesion-level labels, whose decisions are derived like those of the slide model (open diamonds); markers, mean of four PFMs. Right, the gain from showing the model the lesion (ii minus i) and from the region classifier (iii minus ii), means of four PFMs, with the number of PFMs whose 95\% patient-bootstrap interval lies above zero (bold, at least three). Coloured squares, classes compared (grey, all other classes); shaded rows, decisions that involve atypia. Bottom, change in (iii) when a one-patch ring of surrounding tissue is added to each region (range over PFMs and decisions).}
\label{fig:map}
\end{figure*}

To separate what works from what does not, we decomposed the seven-way output into nine clinical decisions (Fig.~\ref{fig:map}b, grey circles). The decisions that involve malignancy performed well, for example malignant versus the rest (AUC 0.94--0.95). In contrast, the decisions that separate atypia from its benign and in situ neighbours performed poorly, from 0.52--0.58 for FEA versus ADH to 0.75--0.79 for benign versus atypical. Only atypical versus malignant stayed high (0.89--0.91). In the format of the pathologist study of Elmore et al.~\citep{elmore_concordance_2015}, the model also under-graded 50--60\% of the DCIS slides, far more than practising pathologists (Supplementary Note S17).

\begin{figure*}[p]
\centering
\includegraphics[width=\textwidth]{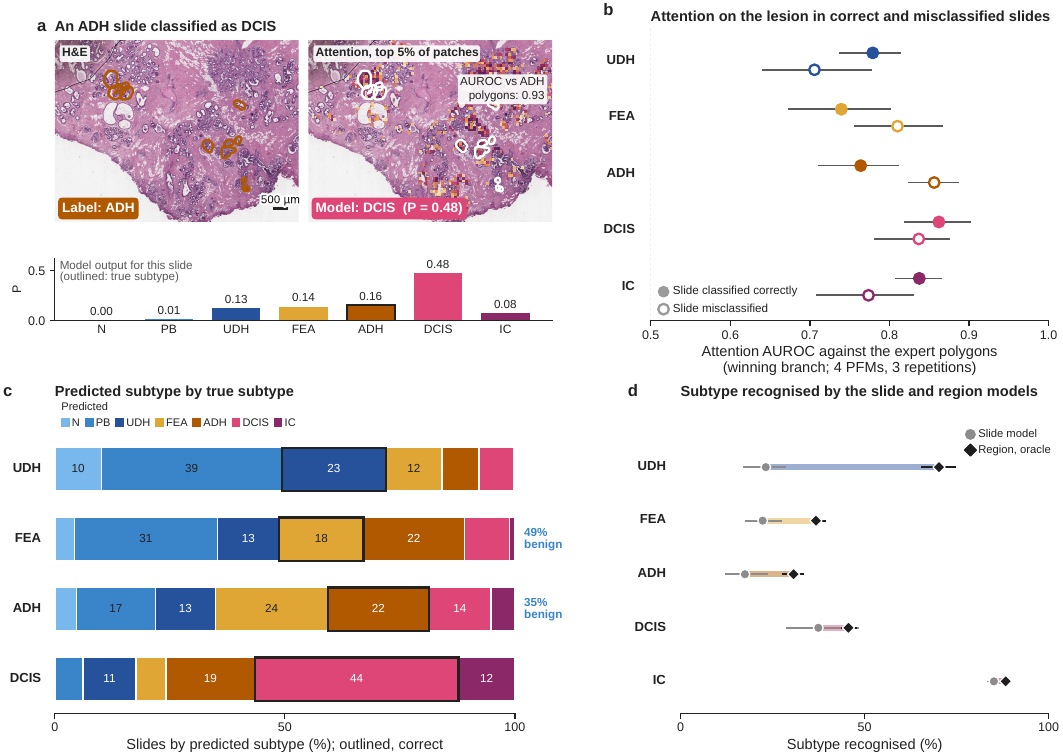}
\caption{\textbf{Attention and classification of the lesion.} \textbf{a}, An ADH slide classified as DCIS (Virchow2, fold model of the first repetition in which the slide is in the test set). Left, H\&E with the expert ADH polygons; right, the 5\% of patches with the highest attention of the winning branch, with the polygons in white; below, the model's seven-way output (outlined, true subtype). The attention AUROC of this slide is the median of misclassified ADH slides for this encoder. Scale bar, 500~$\mu$m. \textbf{b}, AUROC of the attention of the winning branch against the patches inside the expert polygons of the slide's own class, in correctly (filled) and incorrectly (open) classified slides; markers, mean over patients (four PFMs and three repetitions pooled); whiskers, 95\% patient-bootstrap intervals. \textbf{c}, Subtype predicted by the slide model for slides of each true subtype (mean of four PFMs; out-of-fold probabilities averaged over three repetitions); outlined, the correct subtype; right, percentage called benign. \textbf{d}, Percentage of slides whose subtype is recognised by the slide model (circles) and by the seven-way region classifier with oracle localisation (diamonds); markers, mean of four PFMs; whiskers, range.}
\label{fig:found}
\end{figure*}

A decision can fail at two steps. The model may not locate the lesion, or it may locate it and assign it the wrong subtype. We first measured how well attention located the lesion that defines the label. Attention ranked the patches of the lesion above the rest of the slide (AUROC 0.79--0.84 against the expert polygons), and it did so equally well in misclassified and in correctly classified slides (Fig.~\ref{fig:found}b). However, it rarely attended to the lesion first. In 60--73\% of the atypical slides, none of the 16 most attended patches fell inside the lesion, which covered a median of 0.17\% of the slide (Fig.~\ref{fig:found}a; Supplementary Note S14).

If the problem were locating the lesion, giving it to the model should improve the atypical decisions. To test this, we evaluated each decision in three ways on the same slides and folds (Fig.~\ref{fig:map}a). The first evaluation was the slide model on the whole slide, the second the same model given only the patches of the expert polygons of the slide's class, and the third a region classifier trained on those polygons with lesion-level labels. Given only the lesion, the slide model improved benign versus atypical, but not the three boundaries within atypia (Fig.~\ref{fig:map}b). The region classifier, in contrast, improved all four atypical boundaries (benign versus atypical, UDH versus ADH, FEA versus ADH and ADH versus DCIS) by a further $+0.06$ to $+0.29$ in AUC, and localisation accounted for only a median of 0.25 of the total gain. Lesion-level labels alone did not explain the rest. A single seven-way classifier trained on the same regions improved only FEA versus ADH (Fig.~\ref{fig:map}b, open diamonds), and adding the tissue around each region did not help (Supplementary Notes S10 and S12). Locating the lesion more precisely would therefore not be enough. The main problem is the classification of the lesion, and for three of the four atypical boundaries, improving on the lesion-only evaluation required both lesion-level labels and a classifier trained for that boundary.

The errors of the slide model are consistent with this. It classified only 22\% of the ADH slides correctly and called 35\% of them benign (Fig.~\ref{fig:found}c). Even with lesion-level labels, a seven-way region classifier recognised ADH in only 0.28--0.33 of the ADH slides (Fig.~\ref{fig:found}d), although the pairwise classifiers described above separate ADH from each of its neighbours. Predictive uncertainty did not single out the atypical errors either (Supplementary Note S17).

\subsection*{Each atypical boundary fails in line with its clinical definition}

\begin{figure*}[p]
\centering
\includegraphics[width=\textwidth]{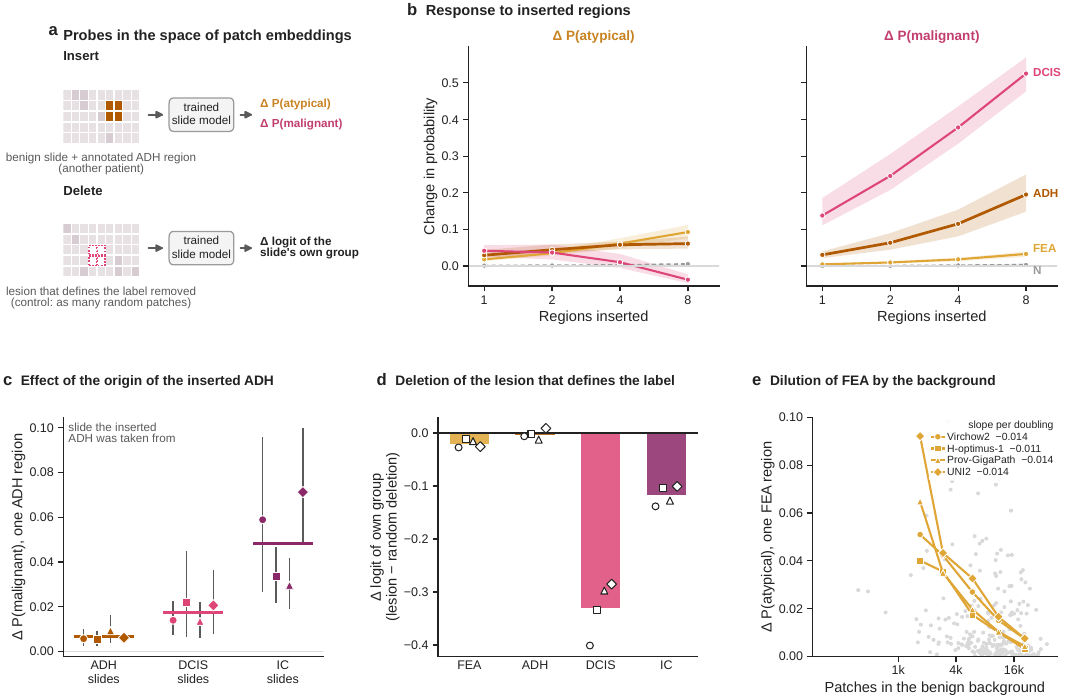}
\caption{\textbf{Causal probes of the slide model.} \textbf{a}, Probes edit the bag of patch embeddings of a test slide and record the change in the trained model's output: insertion of annotated regions taken from another patient into benign slides, and deletion of the lesion that defines the label, with the deletion of as many random patches as control. \textbf{b}, Change in the probability of the atypical (left) and malignant (right) superclass after inserting one to eight regions of each subtype into benign slides; lines, mean of four PFMs; bands, range across PFMs; N, normal tissue (composition control). \textbf{c}, Change in the probability of malignancy after inserting one ADH region, by the label of the slide the region was taken from; markers, individual PFMs; whiskers, 95\% bootstrap intervals over donor slides; horizontal lines, means of four PFMs. \textbf{d}, Change in the logit of the slide's own superclass when the lesion that defines the label is deleted, minus the change when as many random patches are deleted; bars, mean of four PFMs; markers, individual PFMs. \textbf{e}, Dilution: change in the probability of atypia after inserting the same FEA region into benign backgrounds of different size; grey dots, mean per background slide (Virchow2); lines, means within one-doubling bins that contain at least ten background slides; legend, slope per doubling of the background. Marker shapes identify the PFM: circle, Virchow2; square, H-optimus-1; triangle, Prov-GigaPath; diamond, UNI2. Donor regions and background slides are test slides of the same fold, from different patients (15 fold models per PFM).}
\label{fig:probes}
\end{figure*}

Next, we used causal probes to study how the model responds to each type of lesion (Fig.~\ref{fig:probes}a). Inserted ADH regions raised the probability of malignancy more than that of atypia in all four PFMs, and the gap widened with the dose, since eight regions added $+0.20$ to P(malignant) and $+0.06$ to P(atypical) (Fig.~\ref{fig:probes}b). The origin of the ADH mattered. An ADH region taken from an IC slide raised P(malignant) by 0.030--0.071, whereas one from an ADH slide raised it by only 0.005--0.009 (Fig.~\ref{fig:probes}c). This is the same ADH that lies next to DCIS in the region embedding. We then asked whether this reading is learned from the label. Removing from training the 75 malignant slides that contain annotated ADH or FEA did not weaken it in any PFM, so it is not taught by the maximum rule. In addition, deleting the lesion that defines the label lowered the output of DCIS slides but not that of FEA or ADH slides (Fig.~\ref{fig:probes}d), and the effect of an FEA region decreased as the benign background grew (Fig.~\ref{fig:probes}e).

\begin{figure*}[p]
\centering
\includegraphics[width=\textwidth]{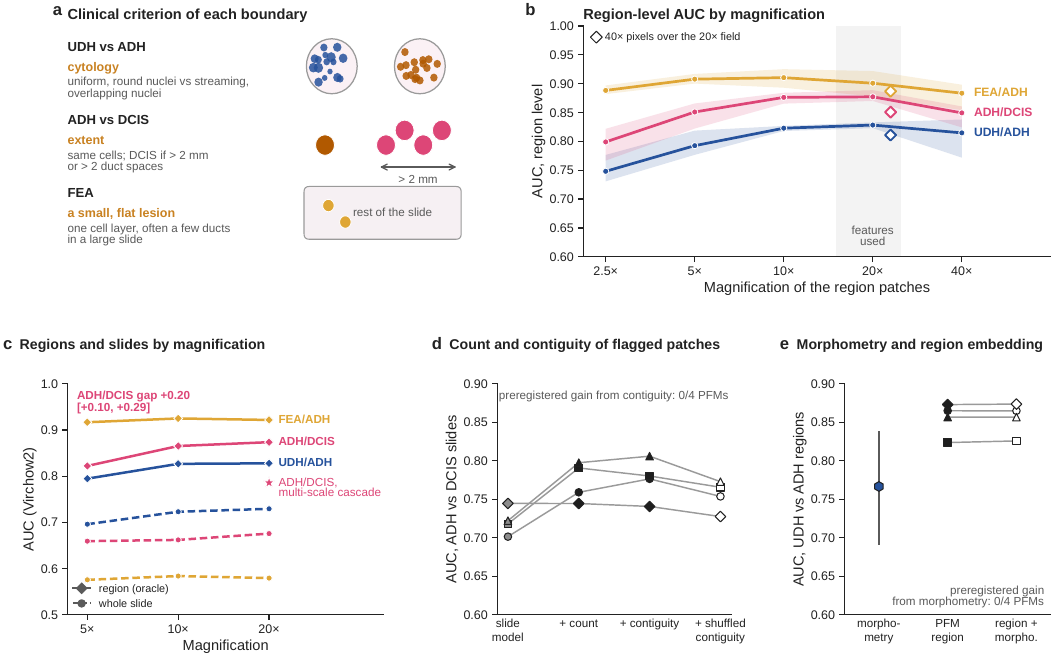}
\caption{\textbf{Magnification, extent and morphometry for each atypical boundary.} \textbf{a}, The three atypical boundaries and the criterion that defines each: cytology (UDH versus ADH), extent (ADH versus DCIS: the same cells, DCIS when larger than about 2~mm or two duct spaces) and a small, flat lesion in a large slide (FEA). \textbf{b}, Region-level AUC (each polygon one sample) when the polygons are read at 2.5$\times$ to 40$\times$ (224-pixel tiles); lines, mean of four PFMs; bands, range across PFMs; open diamonds, the 20$\times$ field read with 40$\times$ pixels; shaded, the magnification used throughout this study. \textbf{c}, The same boundaries at 5$\times$, 10$\times$ and 20$\times$ on annotated regions (region-level AUC; diamonds, solid lines) and on whole slides by the slide model retrained at each magnification (slide-level AUC; circles, dashed lines), Virchow2. Star, ADH/DCIS for the multi-scale cascade, whose patch criteria are trained with the annotated polygons (Methods); text, region--slide gap for ADH/DCIS at 20$\times$ with its 95\% joint patient-bootstrap interval. \textbf{d}, ADH/DCIS AUC of the slide model alone (grey), after adding the number of patches flagged by an out-of-fold ADH patch scorer (+ count), after adding also the size of their largest contiguous cluster (+ contiguity), and with the same cluster size after shuffling the flags over the slide's own patch positions (+ shuffled contiguity; open); first repetition. \textbf{e}, UDH versus ADH on the regions of each slide's own class: nuclear and architectural morphometry alone (hexagon; whisker, 95\% patient-bootstrap interval), the PFM region embedding (filled) and both together (open). Marker shapes in d and e identify the PFM as in Fig.~\ref{fig:probes}.}
\label{fig:boundaries}
\end{figure*}

The three atypical boundaries then failed in different ways, each in line with its clinical definition (Fig.~\ref{fig:boundaries}a). The distinction between UDH and ADH is cytological. On regions, UDH/ADH improved from 2.5$\times$ to 10$\times$ and gained nothing beyond 20$\times$, so the 20$\times$ tiles used throughout do not miss cytological information that these encoders can use (Fig.~\ref{fig:boundaries}b; Supplementary Note S6). Explicit nuclear and architectural morphometry separated UDH from ADH on its own (AUC 0.77 [0.69, 0.84]), but it added nothing to the PFM region embedding (Fig.~\ref{fig:boundaries}e). The cytology that these measures capture therefore seems to be already in the representation.

The distinction between ADH and DCIS is quantitative. At the slide level this decision did not improve with magnification, whereas on regions it was much better at every magnification (gap at 20$\times$, $+0.20$ [$+0.10$, $+0.29$]; Fig.~\ref{fig:boundaries}c). What the slide model lacks is therefore a measure of how much lesion there is. A count of suspicious patches did not provide it, and their contiguity added nothing (Fig.~\ref{fig:boundaries}d). A cascade that measures the area of suspicious tissue explicitly improved ADH versus DCIS by $+0.11$ [$+0.03$, $+0.19$], but its criteria are trained with the annotated polygons, so this gain cannot be attributed to the measurement alone. Finally, FEA was the most separable atypical boundary on regions and the one at which the slide model came closest to chance (0.52--0.58). The dilution measured with the probes may explain this difference.

\subsection*{Triage specificity and the UDH/ADH boundary do not transfer to new sites}

We applied the BRACS models without adaptation to the two external cohorts. In the hospital, each slide contains a single subtype selected and labelled by a pathologist, so the maximum rule does not mix lesions. HISTAI, a public collection from another institution and country~\citep{nechaev_histai_2025}, labels each case with the most severe lesion of its pathology report and was extracted with Virchow2 only (Methods).

\begin{figure*}[p]
\centering
\includegraphics[width=\textwidth]{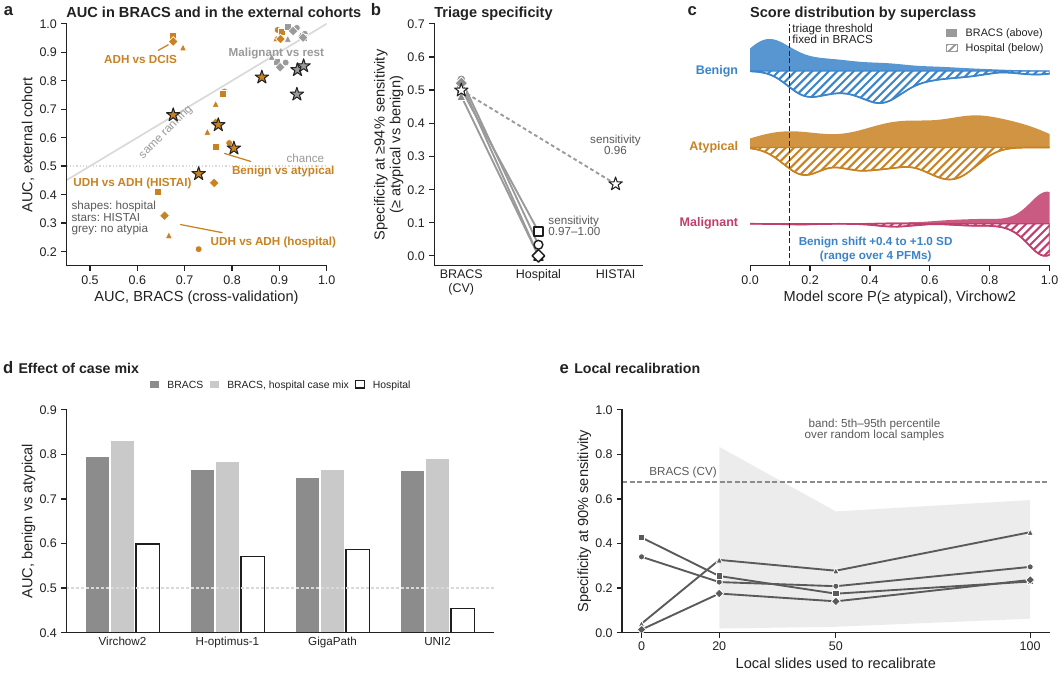}
\caption{\textbf{External validation on pathologist-labelled slides.} \textbf{a}, AUC of each decision in the external cohorts against BRACS: hospital (marker shapes, four PFMs, against BRACS cross-validation; one subtype per slide) and HISTAI (stars, Virchow2, against BRACS slides of the same subtypes; case-level labels); orange, decisions that involve atypia; diagonal, equal AUC. FEA is not evaluated externally; dotted line, chance. \textbf{b}, Specificity for atypia-or-worse versus benign at a threshold fixed in BRACS cross-validation for a sensitivity of at least 94\%, in BRACS, in the hospital and in HISTAI (stars; BRACS reference restricted to the HISTAI subtypes); text, sensitivity reached externally. \textbf{c}, Distribution of the score P($\geq$ atypical) by true superclass, in BRACS (filled, above the line) and in the hospital (hatched, below), Virchow2; dashed line, the triage threshold of b; text, displacement of benign slides in BRACS standard deviations (range over four PFMs). \textbf{d}, Benign versus atypical AUC in BRACS, in BRACS reweighted to the hospital's case mix, and in the hospital. \textbf{e}, Specificity at 90\% sensitivity in the hospital after setting the threshold on $k$ local slides and testing on the rest ($k=0$, the BRACS threshold); lines, mean over 200 random draws for each PFM; band, 5th--95th percentile over draws (mean of the four PFMs); dashed line, specificity at 90\% sensitivity in BRACS cross-validation. Marker shapes in a, b and e identify the PFM as in Fig.~\ref{fig:probes}; a, b, d and e use patient-grouped cross-validation with three repetitions, and the external cohorts are scored by the mean of the five fold models of each repetition.}
\label{fig:deploy}
\end{figure*}

In the hospital, the decisions that involve malignancy kept their ranking (Fig.~\ref{fig:deploy}a), and ADH versus DCIS reached 0.92--0.96 on 11 ADH and 15 single-subtype DCIS slides, well above the 0.67--0.70 of BRACS. In HISTAI, malignant versus the rest fell from 0.95 on matched BRACS slides to 0.85. The atypical decisions weakened in both cohorts. UDH versus ADH failed completely. In the hospital its ranking was reversed (AUC 0.21--0.41 on 8 UDH and 11 ADH slides), because UDH slides received a higher probability of atypia or worse than ADH slides, and in HISTAI it fell to 0.47, the level of chance. Because the hospital labels are pure, this failure cannot be attributed to the label.

A triage threshold fixed in cross-validation for a sensitivity of at least 94\% to atypia or worse kept its sensitivity in both cohorts. However, its specificity fell from 0.48--0.53 to 0.00--0.07 in the hospital and from 0.50 to 0.22 in HISTAI (Fig.~\ref{fig:deploy}b). The loss came from the benign slides, which were scored $+0.41$ to $+0.97$ SD closer to lesion than in BRACS (Fig.~\ref{fig:deploy}c). Neither prevalence nor case mix explained this shift. Reweighting BRACS to the composition of the hospital predicts a benign/atypical AUC of 0.77--0.83, whereas 0.45--0.60 was observed (Fig.~\ref{fig:deploy}d).

To identify the source of the shift, we trained a domain classifier to tell single patches of the hospital from those of BRACS. It did so with an AUC of 1.00, also on fat, a tissue with no diagnostic content. The classifier therefore identified the centre from differences in image acquisition, as Badgeley et al.\ observed for the scanner in radiographs~\citep{badgeley_hip_2019} and as was observed earlier in public pancreatic cancer cohorts, where the features of a weakly supervised model separated the datasets rather than the classes~\citep{carrillo_pdac_2023}. The shift of the benign slides pointed along this domain direction (cosine 0.98).

Removing these differences restored only part of the specificity. Matching the mean and spread of each feature to BRACS, or erasing the domain direction with LEACE~\citep{belrose_leace_2023} and retraining, raised the specificity by $+0.02$ to $+0.05$ in three PFMs. In UNI2 the gain was larger, but the sensitivity fell to 0.87 and 0.89, respectively. Colour normalisation of the images~\citep{reinhard_color_2001} met the end point in only 2/4 PFMs, and in HISTAI it lowered the specificity from 0.22 to 0.15. A threshold recalibrated on 100 local slides did not restore the operating point either (Fig.~\ref{fig:deploy}e). These and the other corrections are detailed in Supplementary Note S5.

All these findings were reproduced with two other aggregators, ABMIL~\citep{ilse_attention-based_2018} and MambaMIL~\citep{yang_mambamil_2024}, each in at least 3/4 PFMs, and neither closed the atypical gap (Supplementary Note S11). Only one finding depended on the branch per subtype. Without it, lower-grade regions inserted into higher-grade slides lowered their predicted grade, which the maximum rule does not allow.

\section*{Discussion}

In this study, we analysed at which steps slide-level classifiers built on PFMs fail on atypia. First, most of the annotated atypia does not appear in the slide labels, because the maximum rule reports only the most severe lesion. Second, the information is still present in the annotated regions, since classifiers trained on the expert polygons with lesion-level labels, one per boundary, separated every atypical boundary in all four encoders. Third, the slide model fails mainly when it has to classify the lesion, and less when it has to locate it. Its attention ranked the lesion above most of the slide, although rarely first, and showing the model only the lesion explained about a quarter of the improvement obtained by the region classifier. For three of the four atypical boundaries, even lesion-level labels improved on the lesion-only evaluation only when each boundary was learned separately.

The hospital cohort helps to separate the label from the model. On its single-subtype slides, ADH versus DCIS was ranked far better than in BRACS, and in BRACS itself the same decision tended to improve when only the DCIS slides without annotated atypia were used (exploratory). Part of the poor internal result on this boundary is therefore probably an effect of the mixed labels, at least in evaluation. On the other hand, UDH versus ADH was reversed on the same pure slides, so this failure belongs to the model and to the change of site, not to the label.

The way in which each boundary failed followed its clinical definition. For UDH versus ADH, the definition is cytological and the evidence is local, and explicit morphometry added nothing to the encoders. For ADH versus DCIS, the definition is quantitative. The evidence therefore has to be aggregated as an amount, which neither an attention-weighted average nor a count of suspicious patches provides. For FEA, the lesion is small and flat, and its signal is diluted by the rest of the slide. Approaches that analyse the structure of the whole slide, rather than single fields, have done better with these lesions. Mercan et al.\ showed that features of ductal architecture separated DCIS from atypia more sensitively than practising pathologists~\citep{mercan_assessment_2019}, and Huang et al.\ predicted the upgrade of ADH and DCIS at excision with a model that measures lesion extent explicitly~\citep{huang_deep-learning_2024}.

A second question is why inserted ADH raises the probability of carcinoma. Most of the annotated ADH lies in carcinoma slides, where it forms a second mode next to DCIS in the frozen region embeddings, and this ADH raised the probability of malignancy the most. Removing these slides from training did not weaken the effect, so it is not learned from the maximum rule but is already present in how the encoders represent these regions. Two recent classifiers, one of them in clinical use, make the same grouping explicitly~\citep{sandbank_validation_2022, yoo_trpv4_2026}. ADH next to carcinoma may be morphologically closer to DCIS as a field effect, which would be consistent with the parallel displacement of benign tissue in the same slides and with the molecular alterations that ADH shares with synchronous low-grade carcinoma~\citep{kader_adh_2018, collins_precursor_2018}. A similar situation was described for TP53 in prostate cancer, where part of the false positives of a slide-level model carried a different alteration that produced the same histological phenotype~\citep{pizurica_tp53_2023}. Alternatively, some of these regions may be parts of DCIS lesions annotated as atypia. The extent criterion did not support this second explanation, but it does not capture cytology, and polygons cover only part of a lesion (only 0.3\% of the DCIS polygons exceed 2~mm), so this question remains open.

These results have direct consequences for deployment. Within its training distribution, a slide-level PFM classifier could be used for triage, flagging carcinoma and ruling out part of the benign slides, but atypia and small DCIS foci must remain with the pathologist. At a new site, neither use can be applied without changes. The triage threshold lost its specificity in both external cohorts, the ranking of malignancy was maintained in only one, and the cytological boundary between UDH and ADH did not transfer to either. In the hospital, one invasive carcinoma was classified as benign by all four encoders with each of the three aggregators, although the triage threshold flagged it with three of the four encoders in most repetitions. The model therefore cannot replace the diagnosis, even for carcinoma.

Drops in performance at external sites are common for these lesions. Dong et al.\ observed a drop in UDH versus DCIS from an AUC of 0.95 to 0.86 when their nuclear classifier was tested in another hospital~\citep{dong_computational_2014}, and across 36 Dutch laboratories the proportion of DCIS graded as high ranged from 30\% to 73\%~\citep{vandooijeweert_grading_2019}, so the reference itself varies between centres. A new site therefore needs local validation, and not only a new threshold.

Correcting this shift was difficult. Removing the differences between centres from the features improved the specificity only partly, and colour normalisation did not improve it. This is consistent with encoders pre-trained on slides from many centres~\citep{zimmermann_virchow2_2024, chen_towards_2024, xu_whole-slide_2024}, whose training data may already cover most of the variation in staining. However, a domain classifier identified the centre from the features with an AUC of 1.00, even in fat. The differences that shift the benign slides are therefore probably not limited to colour, and they include other properties of the acquisition that pre-training did not remove, as also reported for PFMs on other tasks~\citep{komen_towards_2026}.

Our results also have implications for how these models are evaluated. Attention maps are not evidence of a correct diagnosis, because they rank the lesion as high when the class is wrong, and their most attended patches usually lie outside it. Uncertainty cannot tell which atypical calls to trust either. We therefore recommend that studies report AUC or a proper scoring rule per boundary, together with per-class atypical results, rather than the official-split F1. Across four training recipes, this F1 ranged from 0.38--0.54 while the atypical AUC stayed within 0.79--0.83, because it depends on how many slides a model calls atypical (Supplementary Note S17). To make this easier, we release the patient-grouped folds, the per-boundary evaluation, the probes and the case-level labels derived for HISTAI.

Several components of this analysis are not new. The maximum rule is the dataset's own definition~\citep{brancati_bracs_2022}, the probes follow Kaczmarzyk et al.~\citep{kaczmarzyk_explainable_2024}, PFMs are known to encode the acquisition centre~\citep{komen_towards_2026}, and pathologists, handcrafted features and convolutional networks have all had difficulties with intermediate lesions~\citep{elmore_concordance_2015, mercan_assessment_2019, gecer2018}. Previous work classified the lesions themselves~\citep{dong_computational_2014, wetstein_dcis_2021} or compared whole-slide systems with pathologists~\citep{mercan_assessment_2019, gecer2018}. In contrast, this study shows at which step a slide-level classifier fails, boundary by boundary, with preregistered tests in four encoders, three aggregators and two external cohorts. It also explains why common solutions have not improved the results. An ordinal loss designed for the atypical continuum did not differ from cross-entropy over 30 paired seeds, and neither alternative aggregators, top-$k$ attention nor ordinal threshold attention changed atypical discrimination (Supplementary Note S3). None of them measures extent or prevents small lesions from being diluted, which is what the clinical definitions require.

This study has several limitations. The hospital cohort contains only 15 atypical slides (11 ADH, one FEA and three without a subtype) and 8 UDH, so its UDH/ADH result is based on only 19 slides. Its slides were selected to contain a single subtype, and such cases may be clearer than those of routine practice, which could make its rankings optimistic. HISTAI has only 14 ADH cases and was extracted with a single encoder, and because its labels come from case reports, a reported lesion may be absent from the digitised slides. The expert polygons of BRACS are representative rather than exhaustive, so a DCIS slide without annotated ADH may still contain it, and the region-level results give an upper limit of what can be obtained, not the performance of a deployable model. The lesion-only evaluation gives the slide model bags far smaller than those it was trained on, which is why we compared rankings rather than probabilities. Finally, all training used BRACS and H\&E with frozen encoders, and our morphometric measures do not cover chromatin texture, which separated UDH, ADH and low-grade DCIS better than nuclear size and shape in an earlier study~\citep{yamada_nucleic_2016}.

In conclusion, the poor results of slide-level classifiers on atypia are not explained by the image content alone. Most annotated atypia is not reflected in the slide labels, the annotated regions contain enough information to separate the atypical lesions when each boundary is learned from lesion-level labels, and the slide model fails mainly when it has to classify the lesion, even when it is shown where the lesion is. We therefore expect progress on atypia to come less from larger encoders than from aggregation methods that measure what the clinical definition measures, from labels that record more than the most severe lesion, and from evaluations that report what transfers across sites.

\section*{Methods}

\subsection*{Datasets}
\textbf{BRACS.} All primary experiments use the public BReAst Carcinoma Subtyping (BRACS) dataset~\citep{brancati_bracs_2022}: 547 WSIs from 189 patients scanned at 40$\times$ (0.25~$\mu$m/pixel) on Leica Aperio scanners. BRACS provides seven subtypes, namely Normal (N), Pathologically Benign (PB), Usual Ductal Hyperplasia (UDH), Flat Epithelial Atypia (FEA), Atypical Ductal Hyperplasia (ADH), Ductal Carcinoma In Situ (DCIS) and Invasive Carcinoma (IC), which are grouped into three superclasses, Benign (N, PB, UDH), Atypical (FEA, ADH) and Malignant (DCIS, IC). Normal tissue is kept within the Benign superclass, as defined by the dataset authors and used throughout the literature. The slide label is defined by the dataset authors as the most aggressive subtype annotated in the slide (the maximum rule). 387 slides carry region annotations (4{,}540 polygons) that the authors describe as representative rather than exhaustive. We used them only for analysis. The dataset contains 89 atypical slides (41 FEA, 48 ADH).

\textbf{External hospital cohort.} An independent cohort of 118 WSIs from the Hospital Universitario Clínico San Cecilio (Granada, Spain) was digitised under routine conditions on Philips Ultra Fast Scanners (composition: IC 38, PB 31, DCIS 15, N 11, ADH 11, UDH 8, atypical without subtype 3, FEA 1). Models were applied without fine-tuning or domain adaptation. Each slide was selected by the reference pathologist to contain a single subtype, and its subtype was taken from the pathologist's original diagnostic record, so the maximum rule does not mix lesions in this cohort. The cohort contains a single FEA, so boundaries involving FEA are not evaluated in it.

\textbf{HISTAI-breast.} The second external cohort comes from the breast collection of HISTAI~\citep{nechaev_histai_2025}, a public archive (CC BY-NC 4.0) that provides the pathology report of each case in English translation. The scanner is not reported and the files carry no usable resolution metadata, so the slides were read at the native magnification of 20$\times$ stated by the archive. We mapped every report to the BRACS subtypes with a rule-based reader, which agreed with an independent reading on 77 of 79 cases, reviewed every ADH case by hand, and labelled each case with the most severe lesion of its report, the maximum rule applied at the level of the case. The cohort comprises every case with H\&E slides whose most severe lesion was PB, UDH, ADH or DCIS, together with the IC cases whose report also grades the accompanying DCIS (selected for a separate study): 517 cases with 581 slides from 504 patient groups, where cases that share a report form one group. Normal tissue and FEA are almost never the most severe lesion in this archive. Because the label belongs to the case, each case was scored by the mean output over its slides, and a lesion named in the report, especially a small ADH focus, may be absent from the slides that were digitised. No HISTAI case was used for training, model selection or threshold setting.

\textbf{Ethics.} The retrospective use of whole-slide images from San Cecilio University Clinical Hospital (Granada, Spain) was reviewed and approved by the Research Ethics Committee of the Andalusian Regional Government (approval SICEIA-2025-001185, 27 May 2025). All slides were fully de-identified before analysis, and no protected health information was accessible to the research team. The requirement for individual informed consent was waived by the ethics committee given the retrospective nature of the study and the use of fully anonymised archival material, in accordance with Spanish Organic Law 3/2018 on Personal Data Protection and the EU General Data Protection Regulation (Regulation (EU) 2016/679). All methods were carried out in accordance with relevant guidelines and regulations and with the Declaration of Helsinki. BRACS, HISTAI and SPIDER are publicly available under their original licences; no additional approval was required for their use.

\subsection*{Feature extraction}
Four PFMs were used as frozen feature extractors: Virchow2~\citep{zimmermann_virchow2_2024}, H-optimus-1~\citep{hoptimus1}, UNI2~\citep{chen_towards_2024} and Prov-GigaPath~\citep{xu_whole-slide_2024} (architectures, pre-training scale and preprocessing in Supplementary Note S2). A ResNet50 truncated after its third block serves as a weak reference outside the decision rule (Supplementary Note S7). Tissue was tiled at exactly 20$\times$ (448-pixel tiles read at 40$\times$ and resized to 224 pixels, 0.505~$\mu$m/pixel), which is the training resolution of H-optimus-1 and Prov-GigaPath and lies within that of Virchow2, and each encoder was applied with its official preprocessing. Virchow2 embeddings concatenate the class token and the mean of the patch tokens (2{,}560 dimensions), whereas the other encoders use their pooled output (1{,}536). An earlier extraction that normalised every encoder with the statistics of a single model is reported only in Supplementary Note S2. HISTAI was extracted with Virchow2 only, from 224-pixel tiles at its native 20$\times$, with the same recipe.
\subsection*{Slide-level model and evaluation protocol}
All slide-level results use one gated-attention MIL model with one attention branch per subtype (seven branches, as in CLAM-MB~\citep{lu_data-efficient_2021}; hidden size 512, attention size 256, dropout 0.25) and a seven-way softmax head, trained with cross-entropy (``ce7''). Training used Adam (learning rate $2\times10^{-4}$, weight decay $10^{-4}$), one slide per step, and class-balanced sampling with replacement, inverse to the frequency of the seven-way label. Each training fold held out one seventh of its patients as a validation set, and the state with the lowest validation loss was kept (at most 50 epochs, at least 10, patience 10). This recipe was selected before any analysis by a preregistered rule among four candidates, using the validation RPS7 only (Supplementary Note S1). Superclass and boundary scores were derived from the seven-way output. The nine decision boundaries are malignant versus the rest, benign versus the rest, atypical versus the rest, benign versus atypical, atypical versus malignant, UDH versus ADH, FEA versus ADH, ADH versus DCIS and DCIS versus IC.

Internal evaluation used patient-grouped five-fold cross-validation, stratified by the seven-way label and repeated three times (fold files released with the code), so that every slide was tested once per repetition. The external cohorts were scored by the mean of the five fold models of each repetition. For comparability with the literature we also trained on the official split with ten seeds, under four recipes (seven or three output classes; early stopping on the validation loss or on the validation weighted F1). The primary metric was the ranked probability score over the seven subtypes (RPS7), chosen because it had the smallest minimum detectable effect in this cohort. AUCs per decision boundary were secondary, and F1 is reported descriptively because it depends on prevalence and threshold.

\textbf{Other aggregators.} To test whether the findings depend on the aggregator, we trained two more slide models with the same objective, recipe, folds, repetitions and features: ABMIL~\citep{ilse_attention-based_2018}, with a single, non-gated attention pooling (hidden size 512, attention size 128) and a seven-way head, and MambaMIL~\citep{yang_mambamil_2024}, with the configuration released by its authors for BRACS (two sequence-reordering Mamba layers, reordering rate 5), which reads the bag as a sequence in serpentine order and pools it with attention. MambaMIL ran with the authors' CUDA kernels in single precision. The findings to replicate and their decision rules, which were the same as those of the original hypotheses (at least three of four PFMs), were fixed before either model was trained (Supplementary Note S11).

\subsection*{Region-level analyses with oracle localisation}
To ask what can be recovered once the lesion is localised, we read each decision in three ways on the same annotated slides, repetitions and folds. (i) The slide model on the whole slide. (ii) The fold model in which the slide is a test slide, without retraining, applied to a bag that holds only the patches inside the expert polygons of the slide's own class. This evaluation changes where the model looks and keeps its slide-level supervision. Because such bags are far smaller than a slide and lie outside the training distribution, we compared rankings (AUC) and not probabilities. (iii) A region classifier, in which every polygon was represented by the mean embedding of the patches whose reference coordinate lies inside it, an RBF-kernel support vector machine was trained per decision boundary on the regions of the training folds with their own region labels, and the slide score was the mean over the slide's own-class regions. Evaluation (iii) is a ceiling with oracle localisation and lesion-level labels, not a deployable model. For (i) and (ii) the decision score was $P(\text{pos})/[P(\text{pos})+P(\text{neg})]$ of the seven-way output. Before computing (ii), we preregistered that localisation alone recovers a decision if (ii) exceeds (i) with an interval above zero in at least three PFMs (HO1), and the share of the oracle gain that localisation explains, $[(\text{ii})-(\text{i})]/[(\text{iii})-(\text{i})]$, as descriptive (HO2). Because (iii) differs from (ii) both in its labels and in being a dedicated binary classifier, a fourth evaluation (iv), also preregistered, trained a seven-way multinomial logistic regression (standardised features, $C=0.05$, balanced class weights) on the same region embeddings and derived each decision from its averaged seven-way output exactly as for the slide model (HOb: (iv) exceeds (ii) in at least three PFMs). A context variant of (iii) added rings of one and two patches around each polygon, and a seven-way support vector machine with balanced class weights measured whether the region that defines the label is recognised on its own.

\subsection*{Geometry of the atypical bridge}
We studied the geometry of the subtypes in three views, namely the annotated regions (R), the mean of all patches of each slide (S) and the slide embedding of the trained model (M). Rather than reading order from the first principal component, which folds the extremes of curved gradients, we computed the shortest Hamiltonian path through the class centroids, Robinson violations of the centroid distance matrix, and the position $t_\mathrm{AT}$ of the atypical centroid on the benign--malignant segment, with patient-bootstrap intervals. Whether UDH, FEA and ADH lie on a single axis was tested by stochastic order. On one axis, the outer pair must be at least as separable as either inner pair, so the deficit AUC(UDH/ADH) $-$ max[AUC(UDH/FEA), AUC(FEA/ADH)], with out-of-fold pairwise logistic classifiers on the region embeddings, should not be negative (H3: interval below zero).

The distribution of ADH regions on the benign--malignant axis was analysed after the confirmatory results were seen (exploratory). We compared one and two Gaussian components by BIC, took the antimode of the kernel density between the two component means as the cut-off, and described the upper mode by the label of the slide that contains each region. To separate a displacement specific to ADH from a field effect of malignant slides, we compared the shift between malignant and own-label slides of ADH with that of PB regions (difference of differences, patient bootstrap, $B=2000$). Whether upper-mode ADH regions meet the definition of ADH was tested after the bimodality was observed but before any measurement. All eligible regions entered the test, namely ADH of malignant slides above the antimode in at least three PFMs, ADH of malignant slides above it in at most one, and ADH of ADH slides above it in at most one (polygons with at least three patches). For each polygon we measured the maximum Feret diameter, the number of epithelial nests from the morphometry below, and the distance to the nearest DCIS or IC polygon of the same slide. The preregistered contrasts were the proportions with a diameter above 2~mm (upper-mode versus ADH-slide ADH) and within 0.25~mm of a carcinoma polygon (upper- versus lower-mode ADH of malignant slides), with patient-bootstrap intervals ($B=2000$).

\subsection*{Causal probes in the space of patch embeddings}
Probes modified the bag of patch embeddings of a test slide and recorded the change in the model's output~\citep{kaczmarzyk_explainable_2024}. Donor regions came from test slides of the same fold, which the model had never seen, and from a different patient than the background. We used four designs. Design A inserted one to eight annotated regions of a given subtype into benign backgrounds (dose--response), and design B inserted lower-grade regions into higher-grade slides. Design C inserted the same FEA region into backgrounds of different size (dilution), and design D deleted the lesion that defines the label, or an equal number of random unannotated patches as control. Insertions of normal tissue controlled for composition. Hypotheses H5.1--H5.9 and their decision rules were fixed before any probe was run (Supplementary Note S8).

To test whether the reading of ADH as carcinoma is learned from the label (HPb), the slide model was retrained without the malignant training slides that contain ADH or FEA polygons (75 slides in BRACS) and, as control, without the same number of other malignant training slides drawn at random within each fold; validation and test sets were untouched. The H5.2 statistic (mean change in P(malignant) minus change in P(atypical) after inserting four ADH regions) was computed for both arms with identical insertions, and HPb was supported if the paired difference had an interval below zero. Malignant slides without polygons cannot be checked and remain in both arms.

\subsection*{Localisation, attention and uncertainty}
For every test slide with at least one polygon of its own class, we computed the AUROC of attention (all seven branches; the winning branch is the main reading) and of integrated gradients on the features against the patches of the lesion that defines the label, and compared correctly and incorrectly classified slides with a patient bootstrap. Because a high AUROC does not say how much of the attention falls on the lesion, we also report, descriptively, the fraction of the $k$ patches with the highest attention of the winning branch that lie inside the polygons of that lesion ($k=16$ and 64, and $k$ equal to the number of lesion patches), against the fraction of the slide's patches that those polygons cover. Predictive uncertainty was the entropy of the three-superclass output divided by $\log 3$, computed on out-of-fold probabilities averaged over the three repetitions and compared between correctly and incorrectly classified slides of each true superclass (two-sided Mann--Whitney test per PFM). Shared errors counted, for each slide, how many of the four PFMs misclassified its superclass.

\subsection*{Resolution, extent and morphometry}
To ask at which magnification each boundary is visible, every polygon was re-read at eight fixed reference points at 40$\times$, 20$\times$, 10$\times$, 5$\times$ and 2.5$\times$ (224-pixel tiles, official recipe of each PFM), and in a control view that covers the 20$\times$ field with 40$\times$ pixels ($2\times2$ tiles at 40$\times$ around each point). Region-level AUCs were obtained with an RBF-kernel support vector machine under patient-grouped cross-validation (five folds, three repetitions). At the slide level the same slide model was retrained on Virchow2 features at 5$\times$ and 10$\times$, Virchow2 being the only encoder trained across that range. These hypotheses (HR1--HR4) were fixed after an exploratory analysis of the earlier extraction and preregistered as a replication with the official recipe, four PFMs and the new control view.

To ask whether extent is lost in aggregation, a linear patch classifier trained on the annotated patches of the training folds scored every patch for ADH against benign tissue (out of fold). Patches above the 99th percentile of the benign background of the training slides were flagged, and logistic models on the slide model's ADH/DCIS score added the number of flagged patches, the size of their largest 8-connected cluster, or the same cluster size after permuting the flags over the slide's own patch positions, which keeps the count and destroys contiguity (H5.10). To ask whether reading each criterion at its own scale helps, we evaluated a multi-scale cascade that decides from the most to the least severe class, each step reading the magnifications assigned to its criterion and the DCIS step an explicit measure of the area of suspicious tissue. Its patch-level criteria were supervised with the annotated polygons of the training folds, so it is reported only to answer the resolution question, not as a proposed method.

For morphometry, nuclei were segmented with InstanSeg (brightfield nuclei model, version 0.1.1) on 40$\times$ crops of the UDH, FEA, ADH and DCIS polygons, which the model resamples internally to 0.5~$\mu$m/pixel. Nuclei were called epithelial by a rule (mean distance to their five nearest nuclei below 14~$\mu$m, eccentricity below 0.95 and area of at least 10~$\mu$m$^2$) and grouped into epithelial nests (pixels within 8~$\mu$m of an epithelial nucleus, closed with a 6-$\mu$m disc; nests with at least 15 epithelial nuclei). Each nest was described by 19 measures: the variability of nuclear area, staining and spacing; the local orientation order and the fraction of elongated nuclei; the mean nuclear diameter relative to 7.5~$\mu$m; the bimodality of nuclear area and staining; the radial gradients of nuclear area and staining (Spearman correlation with depth in the nest) and the difference in local variability between centre and periphery; nuclear contact; the variability and shape index of the Voronoi cells of the epithelial nuclei; the thickness of the epithelium around lumina, in nuclear diameters; the fraction of the nest occupied by lumina; and the number, radial position and circularity of lumina. Measures were aggregated per region as the mean over nests weighted by their number of epithelial nuclei, and regions with fewer than 30 epithelial nuclei were excluded (29 regions). Measures of lesion size (number of nests, number of epithelial nuclei and nest area) were left out, because size is the axis of extent. On the regions of each slide's own class, standardised logistic models ($C=0.05$) on the region embedding, on the morphometry and on both were compared under the folds of the slide model (Hm2; preregistered, with the measure set amended before any computation).

\subsection*{External cohorts: shift and correction}
In the hospital, we separated four explanations of the external behaviour. These were covariate shift (a domain classifier on slide-level feature statistics), where the shift acts (displacement of scores by true superclass, in BRACS standard deviations), prevalence (expectation--maximisation re-estimation of priors) and case mix (reweighting the internal cohort to the external composition). We also tested an unsupervised affine correction of the scores. Local recalibration set the threshold on $k$ randomly drawn hospital slides and evaluated it on the rest (200 draws per $k$).

HISTAI was analysed under its own preregistration, written before any deployment metric was computed on it, with the case as the unit, PB and UDH as benign, and BRACS out-of-fold predictions restricted to the same subtypes as reference. Four hypotheses replicate the hospital findings: the ranking of malignant against the rest is non-inferior (HX1; lower bound of the difference above $-0.10$), benign cases shift towards lesion more than the rest (HX2; difference in BRACS standard deviations above zero), benign versus atypical degrades (HX3) and the triage specificity at the BRACS threshold falls (HX4). Intervals bootstrap patient groups in HISTAI and patients in BRACS, independently ($B=1000$). The margin of HX1 was widened from 0.05 to 0.10 before the HISTAI predictions existed, after a null simulation in which BRACS played the role of HISTAI showed that the narrower margin failed half the time with identical cohorts. The same simulation showed that a benign shift alone lowers the ranking of malignancy, so HX1 and HX2 are not independent.

To find what identifies the centre, a logistic classifier was trained to tell single patches of the hospital from those of BRACS (Virchow2; out of fold, with folds grouped by slide), and its AUC was computed within each tissue type, with patches typed by a linear classifier trained on the annotated breast patches of SPIDER~\citep{nechaev_spider_2025} (exploratory). We compared the mean shift of benign slides with the domain direction and with the benign-to-lesion direction of BRACS (cosines, against a null from random halves of BRACS benign slides). We then tested five corrections against the uncorrected model on the hospital, with the triage specificity at the BRACS threshold as primary end point (interval of the difference above zero) and non-inferiority of malignant versus the rest (lower bound above $-0.02$), both required in at least three PFMs, or in Virchow2 when only that encoder was available: matching the per-feature mean and standard deviation of the hospital patches to BRACS (Z); matching their full covariance (CORAL~\citep{sun_return_2016}); a domain-adversarial slide model~\citep{ganin_domain_2016} adapted on unlabelled HISTAI slides and evaluated on the unseen hospital (Virchow2 only); linear erasure of the domain from all features with LEACE~\citep{belrose_leace_2023}, followed by retraining; and colour normalisation of the images with Reinhard's transfer in Lab space~\citep{reinhard_color_2001}, using per-slide statistics of the tissue pixels of 64 patches and a fixed reference from 40 BRACS slides, followed by re-extraction of the features with every encoder's official recipe. Z, CORAL and the colour normalisation leave the trained models unchanged. Reinhard normalisation was also applied to HISTAI and evaluated with its own end points. A Macenko normalisation was planned first and discarded, before any extraction, because it failed a visual check.

\subsection*{Statistics and preregistration}
All intervals are 95\% percentile intervals from a patient-level bootstrap with $B=1000$, paired whenever two models are compared on the same slides, and joint over the three repetitions. Every confirmatory hypothesis, its statistic and its decision rule were written in a dated document before the corresponding results existed; a hypothesis counts as supported when it holds in at least three of the four PFMs. Amendments are dated and reported with their reason (Supplementary Note S9). Analyses added after the confirmatory results were seen are labelled exploratory in the text.

\subsection*{Use of large language models}
Claude (Anthropic), used through Claude Code, assisted with writing and reviewing analysis code, producing figures, searching and verifying the literature, and drafting and editing the text. Earlier versions of the text were language-edited with Gemini 3 Pro and Grammarly Pro. The authors designed the study, fixed every hypothesis, reviewed all code, analyses and text, and take full responsibility for the content.

\section*{Data availability}
The BRACS dataset analysed in this study is publicly available at \url{https://www.bracs.icar.cnr.it/}. The HISTAI breast collection~\citep{nechaev_histai_2025} and the SPIDER breast patches~\citep{nechaev_spider_2025} are publicly available from their providers under their own licences; the list of HISTAI cases used here and the labels derived from their reports are released with the code. The external Hospital Universitario Cl\'inico San Cecilio cohort cannot be made publicly available because of patient privacy and institutional data-sharing restrictions; de-identified data may be made available from the corresponding author upon reasonable request, subject to approval by the institutional ethics committee and a formal data transfer agreement. Patient-grouped folds, the preregistration and all analysis outputs are released with the code.

\section*{Code availability}
Code, the patient-grouped folds, the per-boundary evaluation, the probes, the case-level labels derived for HISTAI and the preregistration will be released at \url{https://github.com/NereaHdez/atypia-boundaries} upon acceptance.

\section*{Acknowledgements}
This work was supported by the Grants PID2024-160318OB-I00 and PID2021-128317OB-I00, funded by MICIU/AEI/\-10.13039/\-501100011033 and by ERDF, EU, PCI2023-146016-2 funded by MICIU/AEI/\-10.13039/\-501100011033 and by the European Union, and C-ING-172-UGR23 funded by the Department of University, Research and Innovation of the Andalusian Regional Government (Junta de Andaluc\'ia).

\section*{Author contributions}
N.H.\ conceived the study, designed the methodology, wrote the software, performed the validation, formal analysis and investigation, curated the data, prepared the figures and wrote the original draft. L.J.H.\ reviewed and edited the manuscript. C.R.\ provided the external cohort, curated its data and contributed to the investigation. O.V., I.R.\ and F.M.O.\ supervised the work, acquired funding and reviewed and edited the manuscript. All authors reviewed the manuscript.

\section*{Competing interests}
The authors declare no competing interests.

\section*{Additional information}
\textbf{Supplementary information} accompanies this paper (Supplementary Notes S1--S17).

\includepdf[pages=-]{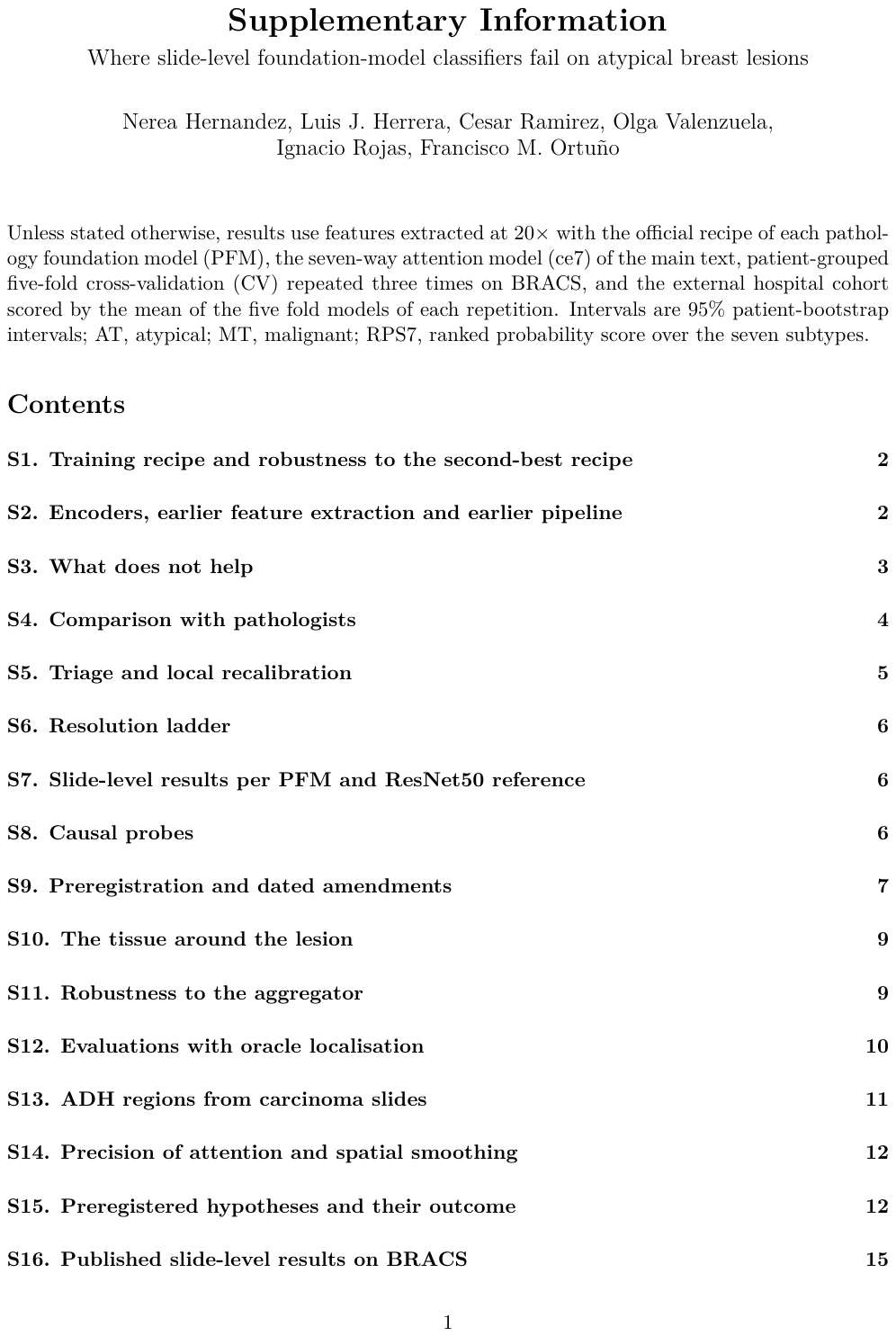}

\end{document}